\documentclass[conference]{IEEEtran}
\IEEEoverridecommandlockouts
\usepackage{booktabs}
\usepackage{multirow}
\usepackage{amsmath,amssymb,amsfonts}
\usepackage{algorithmic}
\usepackage{graphicx}
\usepackage{textcomp}
\usepackage{xcolor}
\usepackage[backend=bibtex,
            style=ieee,
            sorting=none]{biblatex}
\usepackage{url}
\usepackage{hyperref}

\def\BibTeX{{\rm B\kern-.05em{\sc i\kern-.025em b}\kern-.08em
    T\kern-.1667em\lower.7ex\hbox{E}\kern-.125emX}}
\begin{document}

\title{Solar Power driven EV Charging Optimization with Deep Reinforcement Learning\\
\thanks{This project has received funding from the European Union’s Horizon 2020 research and innovation programme under the Marie Skłodowska-Curie grant agreement No 955422.

978-1-6654-6107-8/22/\$31.00 \textcopyright 2022 IEEE
}
}


\author{\IEEEauthorblockN{Stavros Sykiotis}
\IEEEauthorblockA{
\textit{National Technical University of Athens}\\
Athens, Greece\\
stasykiotis@mail.ntua.gr}
\and
\IEEEauthorblockN{Christoforos Menos-Aikateriniadis}
\IEEEauthorblockA{
\textit{National Technical University of Athens}\\
\textit{Intracom S.A. Telecom Solutions, Telco Software Dpt}\\
Athens, Greece \\
}
\and
\IEEEauthorblockN{Anastasios Doulamis}
\IEEEauthorblockA{
\textit{National Technical University of Athens}\\
Athens, Greece \\
}
\and
\IEEEauthorblockN{Nikolaos Doulamis}
\IEEEauthorblockA{
\textit{National Technical University of Athens}\\
Athens, Greece \\
}
\and
\IEEEauthorblockN{ Pavlos S. Georgilakis}
\IEEEauthorblockA{
\textit{National Technical University of Athens}\\
Athens, Greece \\
}
}

\maketitle

\begin{abstract}
Power sector decarbonization plays a vital role in the upcoming energy transition towards a more sustainable future. Decentralized energy resources, such as Electric Vehicles (EV) and solar photovoltaic systems (PV), are continuously integrated in residential power systems, increasing the risk of bottlenecks in power distribution networks. This paper aims to address the challenge of domestic EV charging while prioritizing clean, solar energy consumption. Real Time-of-Use tariffs are treated as a price-based Demand Response (DR) mechanism that can incentivize end-users to optimally shift EV charging load in hours of high solar PV generation with the use of Deep Reinforcement Learning (DRL). Historical measurements from the Pecan Street dataset are analyzed to shape a flexibility potential reward to describe end-user charging preferences. Experimental results show that the proposed DQN EV optimal charging policy is able to reduce electricity bills by an average 11.5\% by achieving an average utilization of solar power 88.4\%. 

\end{abstract}


\section{Introduction}\label{sec:intro}
The transport sector is the end-user sector with the heaviest dependency on fossil fuels, accounting for 37\% of global $CO_2$ emissions according to IEA 2021 Transport report \cite{IEA_transport}. Due to the COVID-19 pandemic, mobility activity and consequently transport $CO_2$ emissions have temporarily decreased. Despite this downward trend, emissions from transport and especially from road passenger vehicles are expected to continuously rise over the next decades. Even if, there is an increasing number of national energy plans targeting net zero emissions by 2050 and renewable energy share is growing bigger in the global electricity generation mix \cite{IEA_Marketreport2022}, high short-term gas prices lead to increased coal-fired power generation. Therefore, the challenge of fully decarbonizing vehicle road transport is growing bigger, especially in the short-term where EV technology is under an evolutionary phase and fuel economy regulations are still under development \cite{IEA_transport}.

Local energy flexibility markets can contribute to the aforementioned direction with the exploitation of self-consumed green energy, providing added value not only to the EV owners, but also to the distribution network operators. Controlling residential EV charging during the day with the design of DR schemes can optimally schedule EV charging load so that end-user electricity costs and carbon footprint are being minimized. The introduction of proper incentives can stimulate end-users to shift their electricity consumption from on-peak hours to low-peak periods and, in that way, tackle network-related issues that may rise.

Artificial Intelligence (AI) can unlock the flexibility potential of low voltage power grids, utilizing smart meter readings to identify domestic appliances signatures from aggregated consumption signals \cite{Sykiotis2022electricity} or schedule and control residential resources to participate in demand response events \cite{Rajasekhar2020Survey,Mabina2021Sustainability}. Many different methods have been used for scheduling and control of residential energy resources with heuristic algorithms, with particle swarm optimization (PSO) and genetic algorithm (GA) to be among the most common, due to their lower computational requirements and the lack of model training needed \cite{Antonopoulos2020Artificial, Menos2022Particle}. However, Reinforcement Learning (RL) has been recently receiving increased attention in the field of demand response applications, since its dynamic character can better integrate uncertainty aspects, such as end-user preferences consideration, in the problem formulation \cite{Liu2020Optimization}. In addition, RL can continuously learn from past experiences in a model-free approach, and consequently increase its performance while in operation \cite{Vazquez2019Reinforcement}. 

The high-dimensional state space in residential resource scheduling and control, especially when considering end-user preferences and PV power self-consumption, has led to an increasing interest in the use of Deep Reinforcement Learning. Solar PV-sourced energy resource scheduling was investigated in \cite{Liu2020Optimization}, where deep Q-learning (DQN) and double deep Q-learning (DDQN) have been compared to PSO on residential resource scheduling. However, EV load scheduling has not been considered in the problem formulation. In \cite{TAI2022}, various residential appliances have been optimally scheduled with the support of solar PV power. Bidirectional power flows have been considered and end-user preferences have been inferred from measured data, but EV charging load has been neglected in the problem design. PV self-consumption optimization has been included also in \cite{LISSA2021}, where Q-learning, a Deep RL method, has been used to optimally schedule domestic space and water heating. In addition, many works have focused on EV charging load scheduling with Deep RL, when considering end-user driving or charging preferences during the day. However, in the majority of reviewed literature \cite{Li2020SafeDQN, Wan2019EV_DRL, Li2020DR, Chis2017PEV}, EV load charging patterns have been modeled with Gaussian probability density functions (stochastic) or considered known (deterministic) without being inferred from historical data. End-user feedback has been included in a DQN algorithm's rewards \cite{Shuvo2022HERS} to minimize electricity bills and user discomfort. However, clean energy prioritization has not been included in the RL environment that models the energy system. In \cite{Ren2022Novel}, EV charging optimization based on historical data-inferred end-user preferences has been conducted,  but renewable power self-consumption has not been prioritized. 

This work proposes a novel framework to optimally charge an EV, while prioritizing PV power consumption and cost minimization, with the use of DQN Reinforcement Learning. Our proposed approach:

\begin{itemize}
    \item Prioritizes solar PV self-consumption for residential EV load scheduling. A solar utilization index is employed to calculate the amount of clean energy that has been self-consumed for EV charging. 
    \item Suggests a flexibility potential index that is introduced in the RL rewards to take into consideration end-user preferences. This index is inferred from analyzing historical consumption data to calculate the average probability for a user to charge the EV at a specific time interval.  
    \item Considers EV battery's technical specifications and daily driving habits through the design of two rewards that are integrated in the RL environment. 
\end{itemize}

This paper is structured as follows. In Section \ref{sec:methodology}, the problem formulation and modelling, both from the energy and the RL perspectives are introduced. In Section \ref{sec:results}, the experimental setup and the evaluation results are presented, whereas in Section \ref{sec:conclusions} the main conclusions of the proposed methodology are drawn.

\section{Methodology} \label{sec:methodology}

The EV load scheduling optimization problem is tackled by examining residential EV owners that have solar PV panels installed on their premises. Therefore, the energy system can be modeled as an individual household connected to the main grid that includes a Battery Electric Vehicle (BEV), an EV home charger, PV panels for solar power generation as well as a residual load comprising of the cumulative power consumption of the remaining domestic house appliances. The real-time EV load scheduling can be formulated as a discrete timestep optimization problem with a 15-minute temporal resolution, where the proposed model aims to optimally distribute the charging load  throughout the day. A 24-hour modeling horizon 

\begin{figure}[ht]
    \centering
    \includegraphics[width=\columnwidth]
    {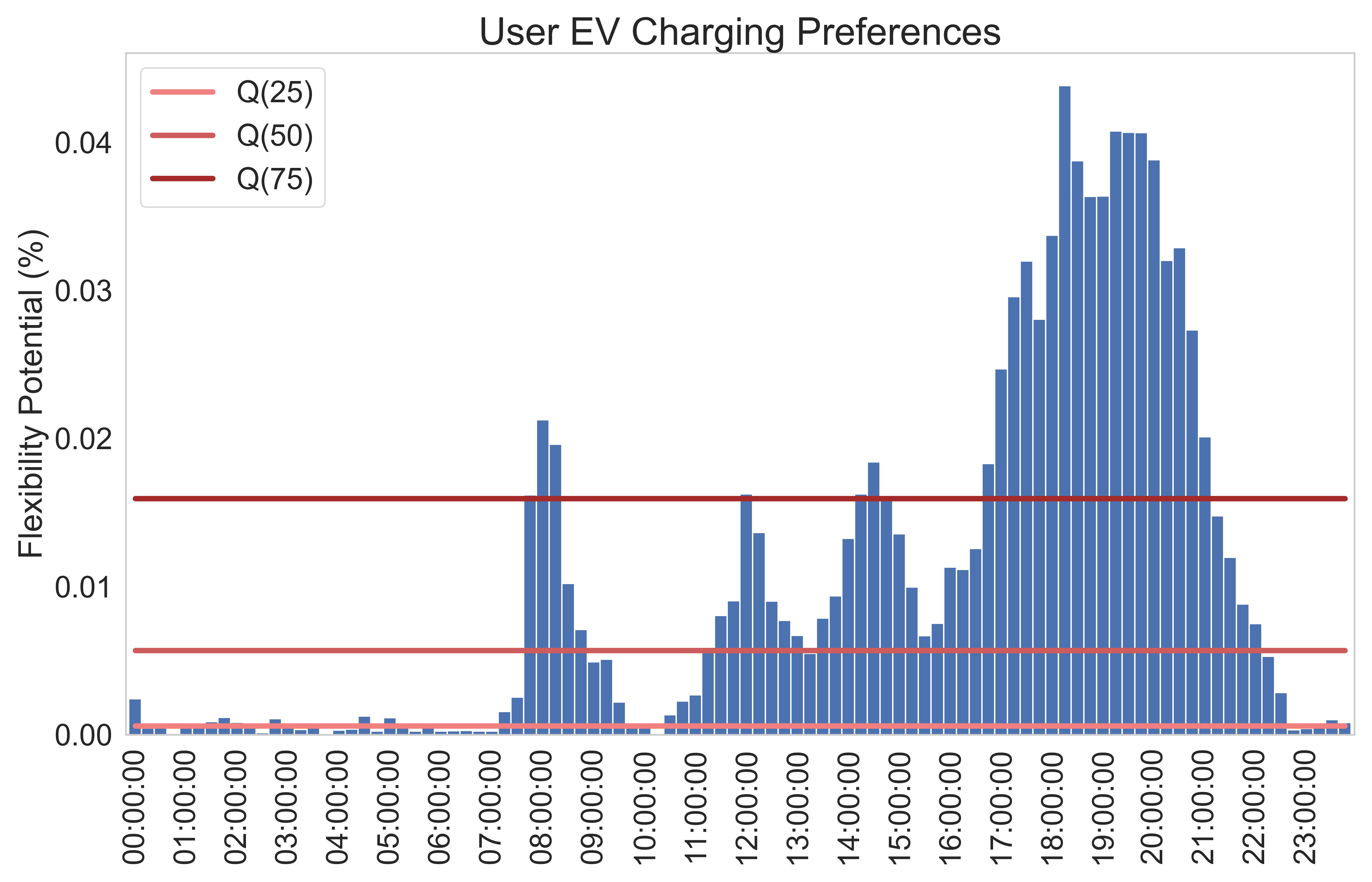}
    \caption{Flexibility potential index for household 4373 in Austin, Texas. The index at each 15-minute time step shows the consistency of EV charging throughout the day (\%).}
    \label{fig:flexindex}
\end{figure}
\noindent
(\textit{T} = 96) is considered and, at each discrete timestep \textit{t}, the model should assess whether charging the EV at the given timeslot would be beneficial for the end-user, as well as estimate the required power amount based on techno-economical criteria without jeopardizing user convenience. In this work, user convenience is formulated through a flexibility potential index which considers the within-the-day EV charging potential. The index is inferred through historical data analysis on real household measurements from Austin, Texas, US from the Pecan Street dataset \cite{Pecan}. The flexibility index profile of an indicative household is illustrated in Figure \ref{fig:flexindex}.

\begin{figure}[ht]
    \centering
    \includegraphics[width = \columnwidth]{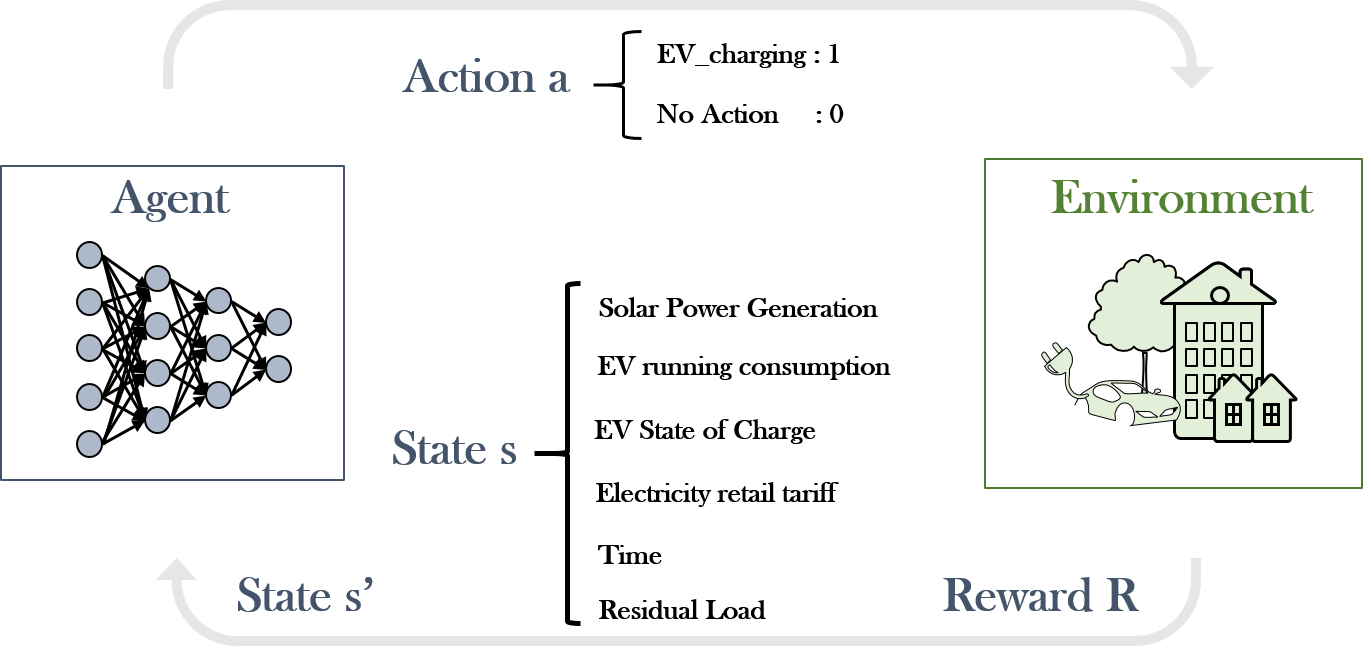}
    \caption{High-level overview of the DQN framework for Demand Response. The agent utilizes a neural network to evaluate an observation state of the environment and choose the corresponding action. In return, it receives a reward based on the optimality of its action.}
    \label{fig:high_level}
\end{figure}

In the context of Deep Reinforcement Learning, the optimization task must be defined as an environment-agent duality. The environment produces observations and the Deep Neural Network driven optimizer (agent) evaluates the given observation and chooses an action, for which it is positively or negatively rewarded. Therefore, the EV load scheduling task is formulated as a Markov Decision Process (MDP) defined by the tuple $(\mathcal{S},\mathcal{A},R,P)$. $\mathcal{S}$ denotes the state space, i.e. the observations that will be evaluated by the agent to choose the optimal action. The set $\mathcal{A} = \{1,0\}$ contains the possible actions that the agent can choose from, meaning that the agent can choose to either charge or not charge the EV, for any given timestep \textit{t}. After the agent chooses an action, the action is evaluated by the environment depending on the defined criteria and will assign the respective reward from the set $\mathcal{R}$. Then, the agent, which in our approach is modeled as a Deep Q-Network (DQN) \cite{mnih2013playing}, receives the next state \textit{s'} and the same process is iterated through the training phase. Finally, set $\mathcal{P}$ contains the probability that action \textit{a} in state \textit{s} at timestep \textit{t} will lead to state \textit{s'} at timestep  \textit{t+1}. This mechanism allows the agent to assess which actions maximize its rewards for a given state \textit{s}, i.e. learn the optimal state-action pairs $(s,a)$. Figure \ref{fig:high_level} visualizes the aforementioned approach. In the following subsections, a detailed description of the structure of each sub-component of the DRL model is presented. 

\subsection{State}
At each timestep \textit{t}, the optimization agent will receive a vector $s_{t} = (\textit{p}_{t}, \textit{P}_{t}^{PV}, \textit{P}_{t}^{non-EV}, {P}_{t,run}^{EV}, \textit{SoC}_{t}, t)$ from the environment. Therefore, each state \textit{s} contains the following information: (1) $p_{t}$ signifies the Time-of-Use (ToU) electricity tariff at the given timepoint, depending on the period of the day (On-Peak, Mid-Peak, Off-Peak); (2) $P_{t}^{PV}$ denotes the solar power generated by the PV panel; (3) $P_{t}^{non-EV}$ describes the residual (non-EV) consumption load; (4) ${P}_{t,run}^{EV}$ indicates the cumulative EV power consumed from the start of the episode (\textit{t=1}) until the current time step \textit{t}; (5) $SoC_{t}$ shows the State of Charge (SoC) of the EV at time step $\textit{t}$ and (6) $\textit{t}$ contains information about the current time step. 

In our approach, episodes are formulated by splitting the data on a daily basis. At the start of each episode, the environment utilizes historical consumption data to calculate the amount of power that was required to charge the end-user's EV ($P_{day}^{EV}$). It is assumed that the consumed power corresponds to a full charging cycle, since the number of charging cycles the BEV has undergone during a single day is uncertain. Therefore, the starting SoC, $SoC_{start}$, is defined as: 

\begin{equation}\label{eq:EV_SoC}
      SoC_{start} =1 - \eta \frac{ P_{day}^{EV}}{4 E_{batt}}
\end{equation}

\noindent
where $\eta$ is the battery pack charging efficiency and $E_{batt}$ is the rated battery capacity, measured in kWh. Through Equation \ref{eq:EV_SoC}, it is ensured that the optimized EV daily power consumption, i.e. after the load shifting procedure, will remain close ($\pm 5\%$) to the original consumption $P_{day}^{EV}$.

\begin{equation}\label{eq:constr_Eday}
  P^{EV}_{day} \in [ 0.95\sum_{t=1}^{T} P^{EV}_t, 1.05\sum_{t=1}^{T} P^{EV}_t  ]
\end{equation}

\subsection{Action}
As previously mentioned, the agent receives a state $s_{t}$ and selects to either charge or not charge the EV. The selected action is therefore:

 \begin{equation}\label{eq:constr_onoff}
  \alpha_t^{EV} = \{1, 0\}, \forall \alpha \in A, \forall t \in T
 \end{equation}
 
\noindent
where $\alpha_t^{EV}=1$ corresponds to charging the BEV and $\alpha_t^{EV}=0$ means that, according to the DQN agent, it is better to not charge and remain idle.

In addition, the agent actions must be constrained by the physical and technical properties of the battery pack:

\begin{equation}\label{eq:EV_SoC_limits}
  {SoC_{min}} \leq SoC_{t} \leq {SoC_{max}}, \forall t \in T
\end{equation}

\begin{equation}\label{eq:EV_updateSoC}
  SoC_{t+1} = SoC_{t} +  \frac{\eta  P_{t}^{EV}}{4  E_{batt}}, \forall t \in T
\end{equation}

\begin{equation}\label{eq:EV_charging_levels}
    P_{t}^{EV} = \begin{cases}
          3.3 kW, & SoC_{min} \leq SoC_{t} \leq 0.9 \\
          1.5 kW, & SoC_{t} > 0.9 \\
         \end{cases}
         , \forall t \in T
\end{equation}
\noindent
where $SoC_{min}$, $SoC_{max}$ denote the minimum and maximum State of Charge. At a given timestep \textit{t}, $SoC_{t}$ expresses the State of Charge of the EV according to Equation \ref{eq:EV_charging_levels}, and $P_{t}^{EV}$ is the charging power consumption.

\subsection{Rewards}

\begin{figure}[t]
    \centering
    \includegraphics[width=\columnwidth]
    {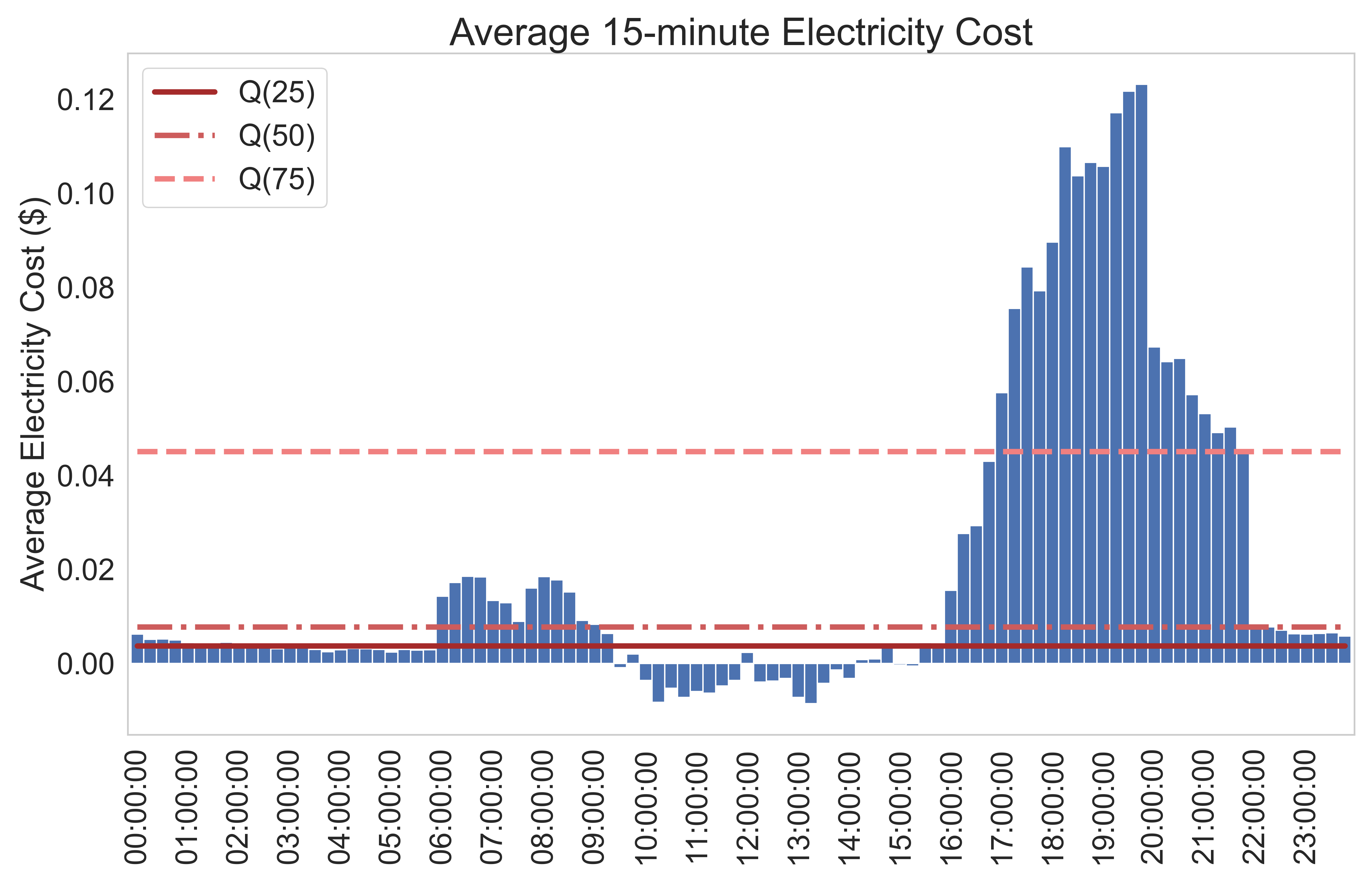}
    \caption{Average Electricity Cost for household \#4373 in Austin, Texas. The cost quantiles are calculated excluding negative 15-minute cost windows.}
    \label{fig:costQx}
\end{figure}

Every new state $s_{t+1}$ is dependent on the current state ($s_t$) and the action that the agent will select ($\alpha_t^{EV}$). Each action will be assigned with a reward from the set $\mathcal{R}$. Therefore, the rewards are the driving force that allow the agent to evaluate whether each action was correct or not and subsequently learn to optimize its decisions. Since the aim of this work is not only to decide the cost-optimal EV charging strategy, but also to utilize green power without violating the battery's operational constraints and the end-user daily habits, a complex multitude of rewards is required. Therefore, we split the total reward into the following sub-rewards: 24-hour power consumption ($r_{1}$), user flexibility potential ($r_{2}$), electricity cost optimization ($r_{3}$) and BEV SoC control ($r_{4}$). Each reward is formulated according to:

\begin{equation}\label{eq:r1}
    r_1 = \begin{cases}
          3,  &  \alpha_t^{EV} = 1 \hspace{2mm} \& \hspace{2mm} P_{t}^{PV} > 0 \hspace{2mm} \& \hspace{2mm} P^{EV}_{t,run} \leq 1.05  P^{EV}_{day} \\
          2,  &  \alpha_t^{EV} = 1 \hspace{2mm} \& \hspace{2mm} P_{t}^{PV} = 0 \hspace{2mm} \& \hspace{2mm} P^{EV}_{t,run} \leq 1.05  P^{EV}_{day} \\
         -10, &  \alpha_t^{EV} = 1 \hspace{2mm} \& \hspace{2mm}  P^{EV}_{t,run} \geq 1.05  P^{EV}_{day} \\
        -0.25, &  \alpha_t^{EV} = 0 \hspace{2mm} \& \hspace{2mm}  P^{EV}_{t,run}  \leq 1.05 P^{EV}_{day} \\
          0,  &  \alpha_t^{EV} = 0 \hspace{2mm} \& \hspace{2mm}  P^{EV}_{t,run} \geq 1.05  P^{EV}_{day} \\
         \end{cases}
\end{equation}
\noindent
From the aforementioned reward it is evident that the agent is strongly rewarded when charging the EV in time steps with solar power generation to promote green energy utilization and self-consumption that reduces electricity costs.

\begin{equation}\label{eq:r2}
    r_2 = \begin{cases}
         -2,  &  \alpha_t^{EV} = 1 \hspace{2mm} \& \hspace{2mm}  U^{flex}_{t}  \leq {Q}_{f}(0.25) \\
         -1, &  \alpha_t^{EV} = 1  \hspace{2mm} \& \hspace{2mm} U^{flex}_{t} \leq {Q}_{f}(0.50) \\
            0, &  \alpha_t^{EV} = 0 \\
          1, &  \alpha_t^{EV} = 1 \hspace{2mm} \& \hspace{2mm}  U^{flex}_{t} \leq {Q}_{f}(0.75) \\
          2, &  \alpha_t^{EV} = 1 \hspace{2mm} \& \hspace{2mm}  U^{flex}_{t} > {Q}_{f}(0.75) \\ 
        \end{cases}
\end{equation}
\noindent
where $U^{flex}_{t}$ expresses the user flexibility potential, i.e. the probability that the user would charge their EV at this timeslot. $Q_{f}(X)$ represents the flexibility potential probability quantile, as illustrated in Figure \ref{fig:flexindex}. $Q_{f}(X)$ is the quantile of BEV charging throughout the day, as obtained from historical data analysis on household \#4373 of the Pecan Street data set \cite{Pecan}. From Equation \ref{eq:r2} it is evident that the agent will receive a higher reward if the probability distribution function of the end-user charging habits is followed. 

\begin{equation}\label{eq:r3}
r_3 = \begin{cases}
    2,  &  \alpha_t^{EV} = 1 \hspace{2mm} \& \hspace{2mm}  C_{t} \leq {Q}_{c}(0.25) \\
    1,  &  \alpha_t^{EV} = 1 \hspace{2mm} \& \hspace{2mm}  C_{t} \leq {Q}_{c}(0.50) \\
    0,  &  \alpha_t^{EV} = 0 \\
    -1, & \alpha_t^{EV} = 1 \hspace{2mm} \& \hspace{2mm}  C_{t} \leq {Q}_{c}(0.75) \\
    -2, & \alpha_t^{EV} = 1 \hspace{2mm} \& \hspace{2mm}  C_{t} > {Q}_{c}(0.75) \\ 
    \end{cases}
\end{equation}
\noindent
where the electricity cost $C_{t}$ is fomrulated according to:
 \begin{equation}\label{eq:obj}
  C_{t} = r_t \cdot (\alpha_t^{EV} \cdot P^{EV}_t + P^{non-EV}_t - P^{PV}_t)
 \end{equation}
\noindent
Figure \ref{fig:costQx} depicts the electricity cost quantiles $Q_{c}(X)$, as well as the daily average electricity cost for house \#4373 of the Pecan Street dataset. For the proper calculation of $Q_{c}(X)$, negative cost periods have been excluded to avoid outlier values in the reward thresholds. Therefore, Equation \ref{eq:r3} clearly shows that a higher cost will lead to a lower reward assignment. 

Finally, sub-reward $r_{4}$ monitors the EV Battery SoC and ensures that the agent is strongly penalized for any action that leads to exceeding the battery capacity ($SoC>100\%$), potentially damaging its physical components.

\begin{equation}\label{eq:r4}
    r_4 = \begin{cases}
          -10, & \alpha_t^{EV} = 1 \hspace{2mm} \& \hspace{2mm} SoC_{t} \geq 1 \\
          0, & otherwise
        \end{cases}
\end{equation}

The total reward \textit{r} that the agent receives is computed as the weighted sum of the previously defined sub-rewards: 

\begin{equation}\label{eq:R_total}
    R = \delta_1 \cdot r_1 + \delta_2 \cdot r_2 + \delta_3 \cdot r_3 + \delta_4 \cdot r_4
\end{equation}
\noindent
All weights are considered equal in this work, and therefore the weight factors $\delta_{k}$ are set to 1.

\section{Results}\label{sec:results}

\subsection{Experimental Setup}

Real measurements for houses in Austin, Texas, provided by the Pecan Street dataset \cite{Pecan}, have been thoroughly analyzed. As described in Section \ref{sec:methodology}, the end user consumption habits have been translated into a flexibility potential reward, and the electricity cost has been calculated using actual residential ToU rates \cite{AustinPrices}, which are shown in Table \ref{tab:tab_prices}. The ToU rates are divided into Off-Peak hours (night and early morning) that correspond to low electricity prices, On-Peak hours (afternnon and evening), where the electricity cost is higher and Mid-Peak hours covering the rest of the day. We assume that the end-user EV is a Nissan Leaf, which has a rated battery capacity of 24 kWh and a Level 2 (AC) slow charger with a charging efficiency of 90.5\% \cite{NissanLeaf}. The evaluation of our approach is conducted on data from house \#4373, as provided by the Pecan dataset \cite{Pecan}. 

\begin{table}[t]
\centering
\caption{2018 summertime, weekdays time-of-use  electricity \\ tariffs of Austin, Texas households}
{\begin{tabular*}{20pc}{@{\extracolsep{\fill}}lll@{}}
\midrule
ToU Period  &   Hours  &    Electricity tariff (\$/kWh)  \\ 
\hline
Off-Peak    &   00:00 - 06:00   &    0.01188  \\
            &   22:00 - 24:00   &             \\
Mid-Peak    &   06:00 - 14:00   &    0.06218  \\ 
            &   20:00 - 22:00   &             \\
On-Peak     &   14:00 - 20:00   &    0.11003  \\
\bottomrule 
\end{tabular*}}{}
\label{tab:tab_prices}
\end{table}
\begin{table}[b]
\centering
\caption{Electricity cost savings on various test set days}
{\begin{tabular*}{20pc}{@{\extracolsep{\fill}}llll@{}}
\toprule
Day              &  Daily EV      & Cost         &  Solar 
\\
                 &  Demand (kW) & Savings (\%) & Utilization (\%)  
\\
\midrule
22/04/2018       &    21.9   &  8.0    & 100        
\\ 
27/06/2018       &    58.5   &  4.2    & 91.8        
\\
08/07/2018       &    78.0   &  11.5   & 76.9
\\ 
12/08/2018       &    10.8   &  5.8    & 69.4
\\ 
13/08/2018       &    41.7   &  5.3    & 96.4
\\
19/08/2018       &    22.2   &  -1.6   & 100
\\
08/09/2018       &    20.7   &  4.6    & 84.1
\\ 
\textbf{Average} &    36.3   &  5.4    & 88.4              \\ 
\bottomrule
\end{tabular*}}{}
\label{tab:tab_results}
\end{table}

\subsection{Results comparison for optimal EV load scheduling}

The proposed DQN agent is trained for 1,000 epochs to learn the optimal cost-reducing EV charging policy by utilizing solar power generation. The model is then evaluated using days that have not been included in the training set. Table \ref{tab:tab_results} presents the cost savings and the total power (kW) sourced by solar PV for EV charging (solar power utilization index) for each day in the test set. On average, our proposed approach achieves average cost savings of 5.4\% with a solar power utilization index of 88.4\%. It can also be observed that the cost savings can reach up to 11.5\%, while in some days the proposed scheme charges the EV completely with PV-generated power.

\begin{figure}[t]
    \centering
    \includegraphics[width = \columnwidth]{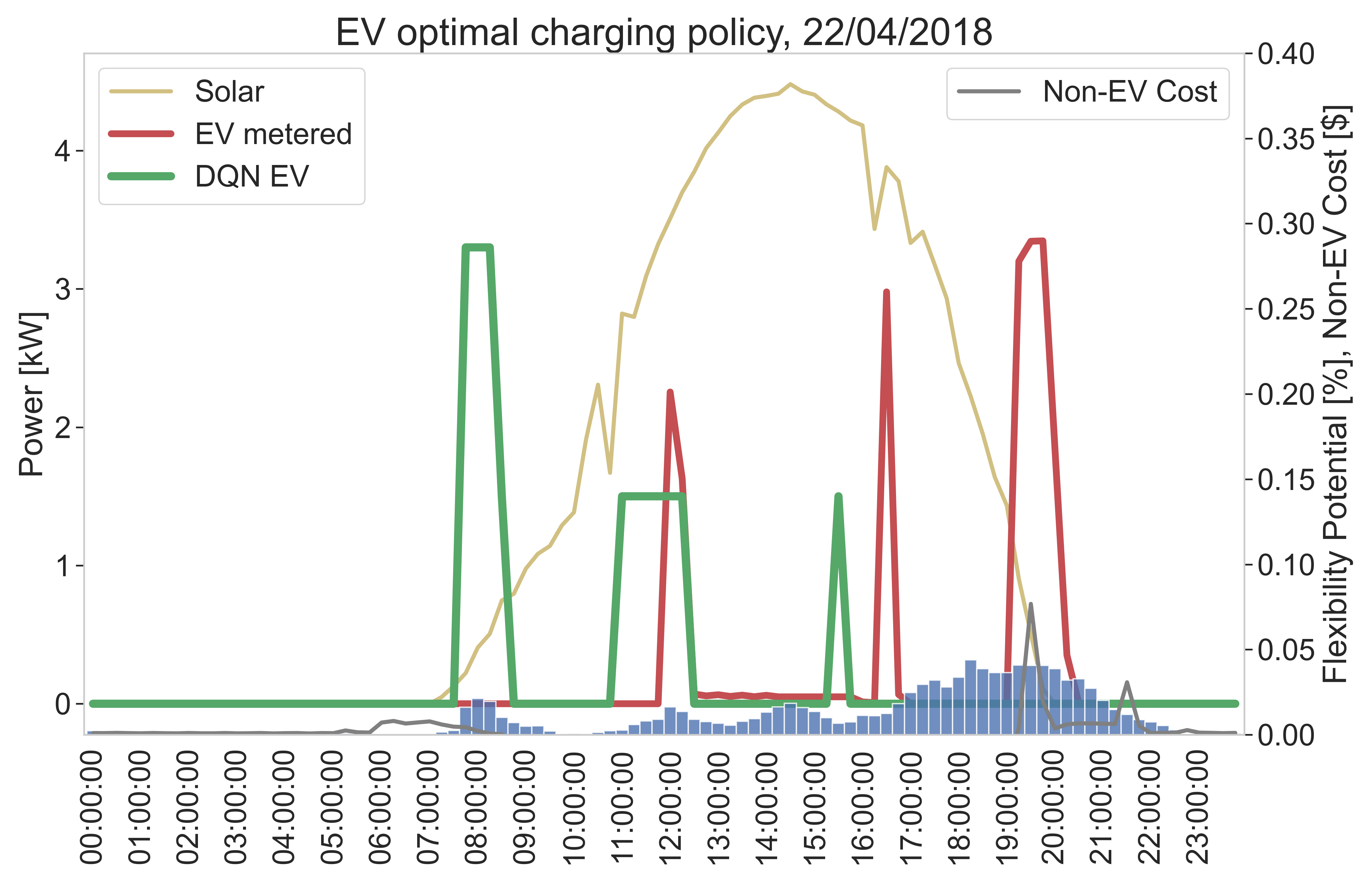}
    \caption{ Comparison between the EV metered and the proposed DQN EV load optimization solutions. The suggested policy matches EV charging load with solar power generation hours without jeopardizing the end-user habits.}
    \label{fig:res_22_4}
\end{figure}
\begin{figure}[t]
    \centering
    \includegraphics[width = \columnwidth]{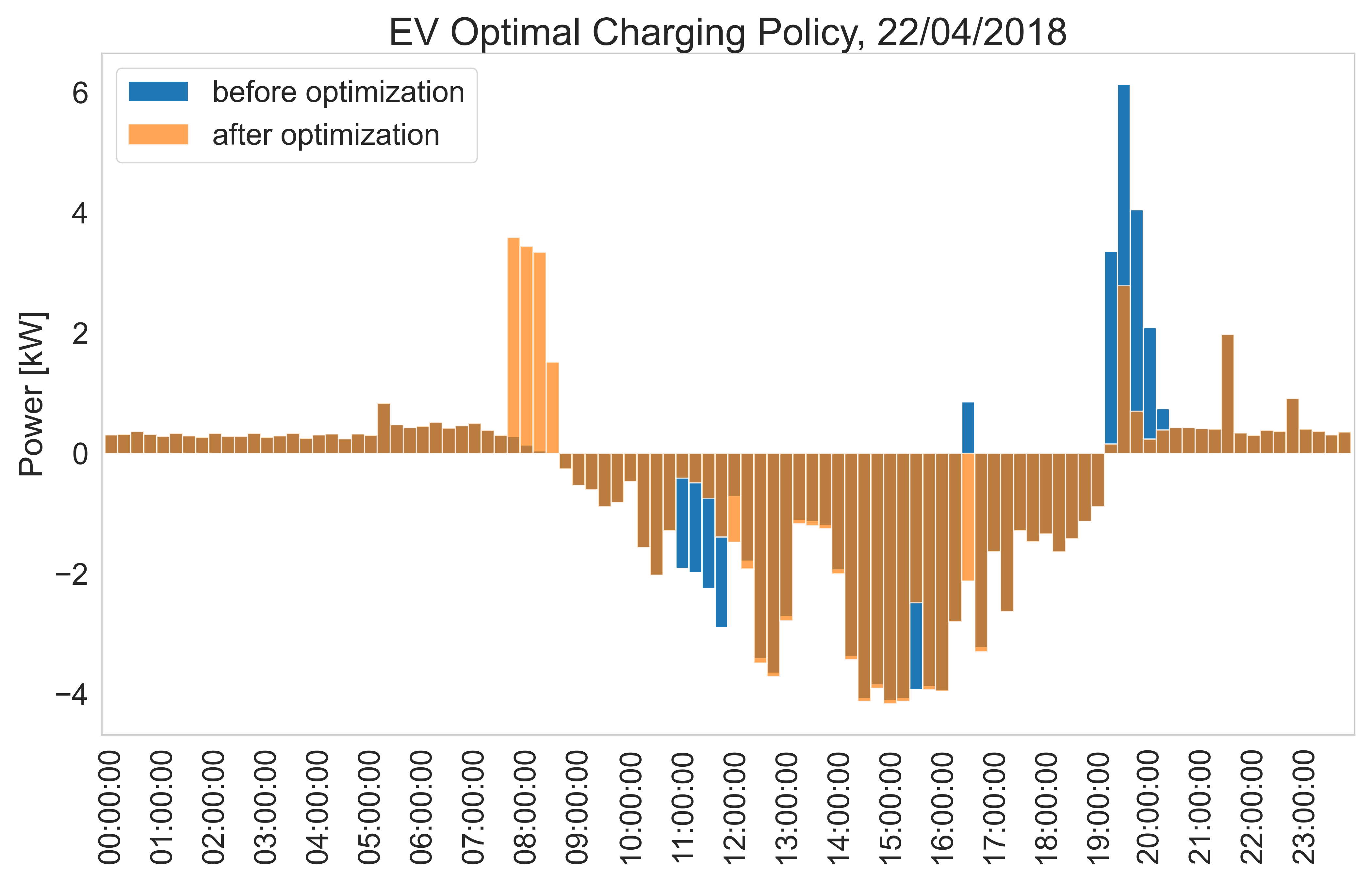}
    \caption{Total grid power consumption comparison between the uncontrolled and controlled solutions. Power consumption during On-Peak hours is being reduced by completely utilizing solar power generation.}
    \label{fig:res_22_4_bar}
\end{figure}

\begin{figure}[t]
    \centering
    \includegraphics[width = \columnwidth]{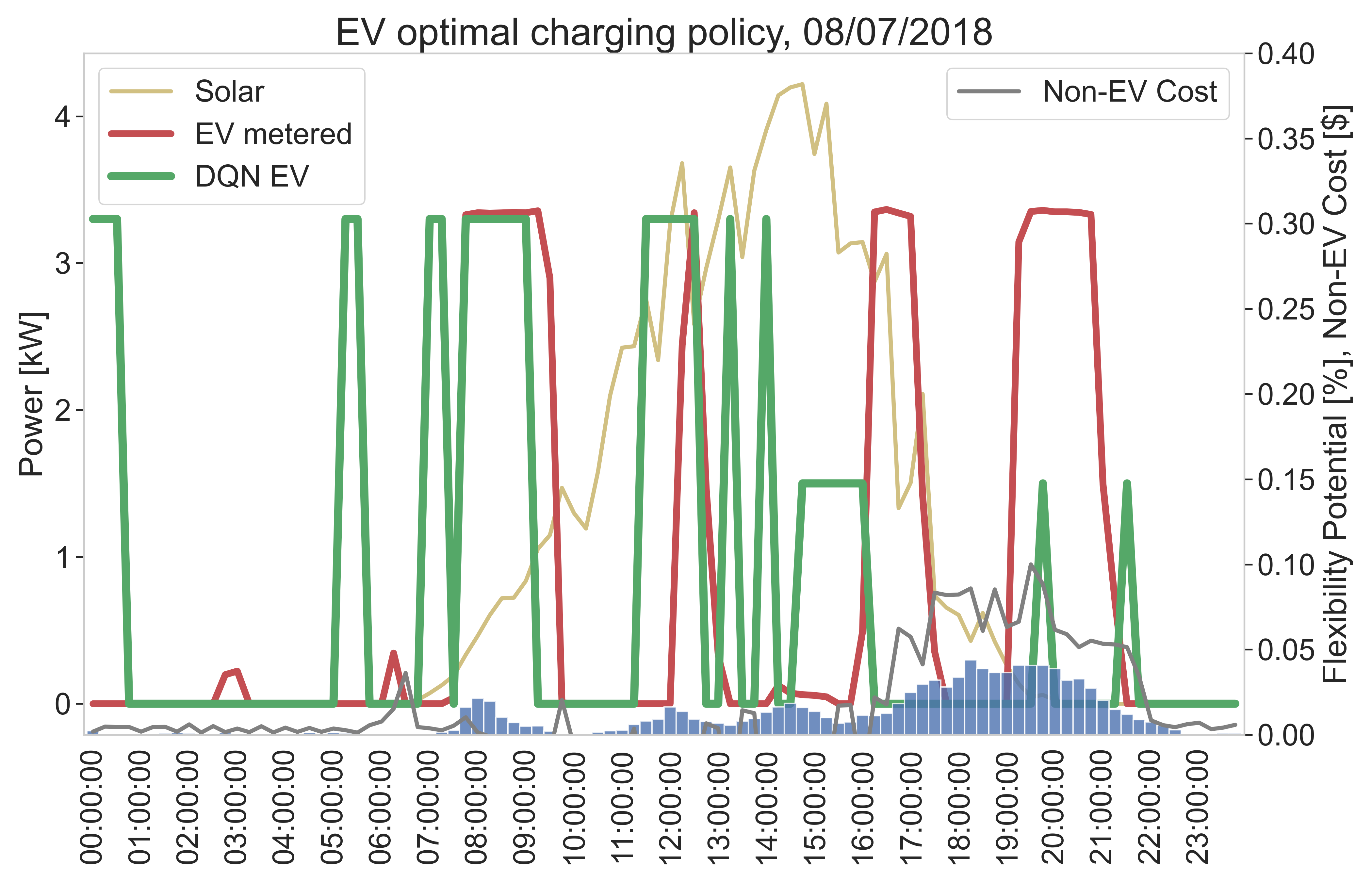}
    \caption{Comparison between the EV metered and the proposed DQN EV load optimization solutions. The suggested charging policy shifts EV load to low-cost periods and achieves an electricity bill reduction of 11.5\%.}
    \label{fig:res_7_8}
\end{figure}

\begin{figure}[t]
    \centering
    \includegraphics[width = \columnwidth]{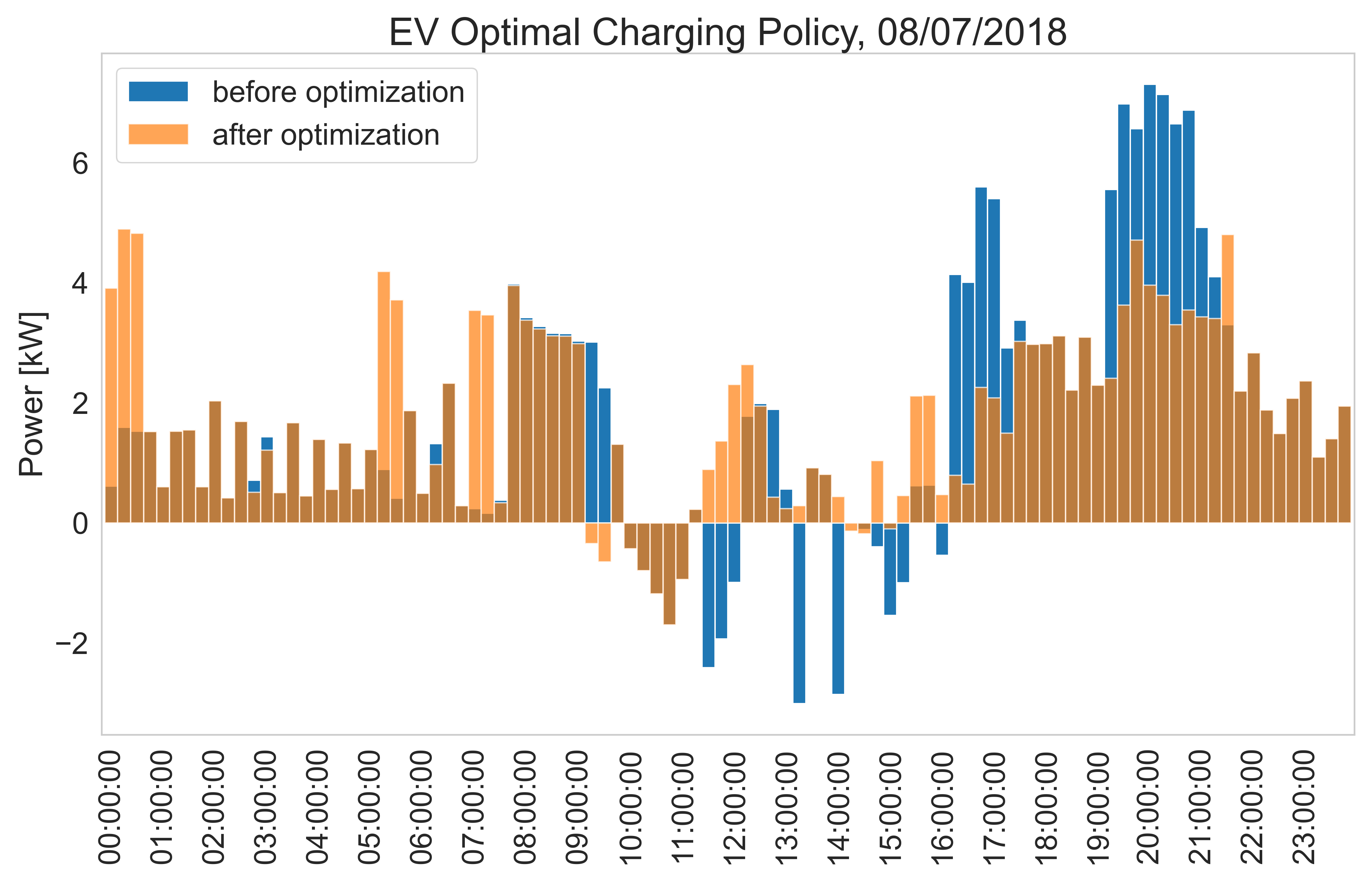}
    \caption{Total grid power consumption comparison between the uncontrolled and controlled solutions. Power consumption during On-Peak hours is being reduced in the suggested DQN EV optimal charging policy.}
    \label{fig:res_7_8_bar}
\end{figure}

The 22/04/2018 test case displays the ability of the DQN EV proposed algorithm to shift completely EV charging in hours with solar PV generation. The solar utilization index reaches 100\% and the electricity cost savings are 8\%, as presented in Table \ref{tab:tab_results}. More specifically, electricity power consumption is being shifted from evening hours to solar power generation periods with low electricity ToU tariffs and high flexibility potential index, as shown in Fig. \ref{fig:res_22_4}. In addition, the proposed algorithm reduced peak power consumption from On-Peak hours, shifting demand to hours with self-consumption, as shown in Fig. \ref{fig:res_22_4_bar}. In the text case of 08/07/2018, higher electricity cost savings can be noticed, reaching up to 11.5 \%. Even if solar utilization on this day remains considerably high (76.9 \%), the EV proposed charging schedule includes hours without any solar PV generation to avoid jeopardizing end user preferences. More specifically, EV charging is being shifted either at night, where electricity tariffs are low, or during hours with high PV power generation, as shown in Fig. \ref{fig:res_7_8}. Similarly to the 22/04/2018 test case, peak power consumption is being reduced as seen in Fig. \ref{fig:res_7_8_bar}.


\section{Conclusions} \label{sec:conclusions}

In this work, a solar PV-driven EV daily load scheduling model has been proposed and modeled with the use of Deep Q-Networks. A set of comprehensive rewards have been introduced to prioritize solar PV self-consumption, end-user EV charging habits and technical/operational constraints in the optimization process. Real household measurements and corresponding ToU tariffs from Austin, Texas, USA have been used to evaluate the efficiency of the proposed DQN model. Experimental results indicate that the suggested EV charging optimization policy can reduce end users' electricity bill by up to 11.5\%. Depending on the daily EV charging demand and the amount of solar PV generation, clean energy utilization for EV charging can reach up to 100\%, paving the way towards road passenger vehicle decarbonisation.

\printbibliography

\end{document}